\documentclass[conference]{IEEEtran}
\IEEEoverridecommandlockouts

\usepackage{cite}
\usepackage{amsmath,amssymb,amsfonts}
\usepackage{graphicx}
\usepackage{textcomp}
\usepackage{xcolor}
\usepackage{amssymb}
\usepackage{booktabs}
\usepackage{subfig}
\usepackage[english]{babel}
\usepackage{xcolor}
\usepackage{makecell}
\usepackage{varwidth}
\usepackage{xcolor}
\usepackage{paralist}
\usepackage{bbm}
\usepackage{multirow}
\usepackage{hyperref}
\usepackage{breqn}
\usepackage{longtable}
\usepackage{mathtools}
\usepackage{algorithm}
\usepackage{algpseudocode}
\usepackage{paralist}
\usepackage{microtype}
\usepackage[detect-all]{siunitx}
\usepackage{cleveref}
\usepackage{lscape} 
\usepackage{upgreek}
\usepackage{graphicx}
\usepackage{booktabs}
\usepackage{multirow}
\usepackage{tcolorbox}

\usepackage{soul,color,xcolor} 
\newcommand{\ours}{ECG-ReGen}

\def\BibTeX{{\rm B\kern-.05em{\sc i\kern-.025em b}\kern-.08em
    T\kern-.1667em\lower.7ex\hbox{E}\kern-.125emX}}
\begin{document}

\title{Electrocardiogram Report Generation and Question Answering via Retrieval-Augmented Self-Supervised Modeling}
\author{
    \IEEEauthorblockN{Jialu Tang\textsuperscript{1}, Tong Xia\textsuperscript{2}, Yuan Lu\textsuperscript{1}, Cecilia Mascolo\textsuperscript{2}, Aaqib Saeed\textsuperscript{1}}
    \IEEEauthorblockA{\textsuperscript{1}Eindhoven University of Technology, Eindhoven, Netherlands}
    \IEEEauthorblockA{\textsuperscript{2}University of Cambridge, United Kingdom}
}

\maketitle

\begin{abstract}
Interpreting electrocardiograms (ECGs) and generating comprehensive reports remain challenging tasks in cardiology, often requiring specialized expertise and significant time investment. To address these critical issues, we propose~\ours, a retrieval-based approach for ECG-to-text report generation and question answering. Our method leverages a self-supervised learning for the ECG encoder, enabling efficient similarity searches and report retrieval. By combining pre-training with dynamic retrieval and Large Language Model (LLM)-based refinement,~\ours~effectively analyzes ECG data and answers related queries, with the potential of improving patient care. Experiments conducted on the PTB-XL and MIMIC-IV-ECG datasets demonstrate superior performance in both in-domain and cross-domain scenarios for report generation. Furthermore, our approach exhibits competitive performance on ECG-QA dataset compared to fully supervised methods when utilizing off-the-shelf LLMs for zero-shot question answering. This approach, effectively combining self-supervised encoder and LLMs, offers a scalable and efficient solution for accurate ECG interpretation, holding significant potential to enhance clinical decision-making.
\end{abstract}


\section{Introduction}
\label{sec:introduction}
Electrocardiograms (ECGs) are non-invasive, cost-effective diagnostic tools that play a crucial role in detecting cardiac arrhythmias in clinical practice. While numerous studies have demonstrated the effectiveness of machine learning models in predicting arrhythmia types from ECGs \cite{minchole2019artificial}, the tasks of interpreting ECGs, generating illustrative reports, and answering patient questions remain largely under-explored \cite{moor2023foundation}. These tasks present significant challenges, as they are time-consuming and require specialized expertise, even for experienced cardiologists. Moreover, they pose unique difficulties for machine learning models, demanding fine-grained feature extraction and the ability to generate cross-modality outputs (i.e., converting signals into coherent textual descriptions).

Recent advances in large language models (LLMs) have sparked considerable interest in their application to medical image interpretation, particularly for generating radiology reports from chest X-rays \cite{chen2020generating,yan2023style}. However, the extension of these approaches to ECG analysis remains largely unexplored, despite ECGs' critical role in cardiology. This gap is significant because ECGs, as biosignals, present unique challenges distinct from static medical images. Converting ECG data into features that LLMs can directly analyze for medical report generation is a complex task, potentially requiring substantial amounts of data, as LLMs were not originally designed for processing time-series biosignals \cite{thirunavukarasu2023large}. Furthermore, concerns persist regarding the efficiency of this process and the models' ability to generalize to new data, such as unseen cardiac conditions or patient populations that differ from those in the training set. These challenges underscore the need for innovative approaches that can effectively bridge the gap between ECG signal processing and the natural language understanding capabilities of LLMs.

In this paper, we propose a novel retrieval-based approach to address the challenges of ECG-to-text report generation and question answering, marking the first application of such a method in this domain. Unlike commonly used task-specific, fully supervised learning approaches~\cite{oh2024ecg,wan2024electrocardiogram}, our method leverages the power of similarity search to establish explicit connections between a given test ECG and examples in an existing dataset. This approach involves interpreting new samples by referencing similar samples in the dataset (e.g.,, training set), making the process more efficient and explainable by design. While such methods have been successfully applied to chest X-ray report generation \cite{endo2021retrieval}, their application to ECGs presents unique challenges. The key to success lies in learning useful representations that aid similarity measures in the feature space, which is particularly challenging for ECGs due to their multi-lead, long time-series nature. Some arrhythmias or cardiac conditions manifest as very subtle changes in the waveforms, making them difficult to detect and requiring significant expertise. Furthermore, ECG reports typically consist of brief and diverse phrases summarizing signal patterns, adding another layer of complexity to the task. 

\begin{figure*}[t]
\centering
\includegraphics[width=\textwidth]{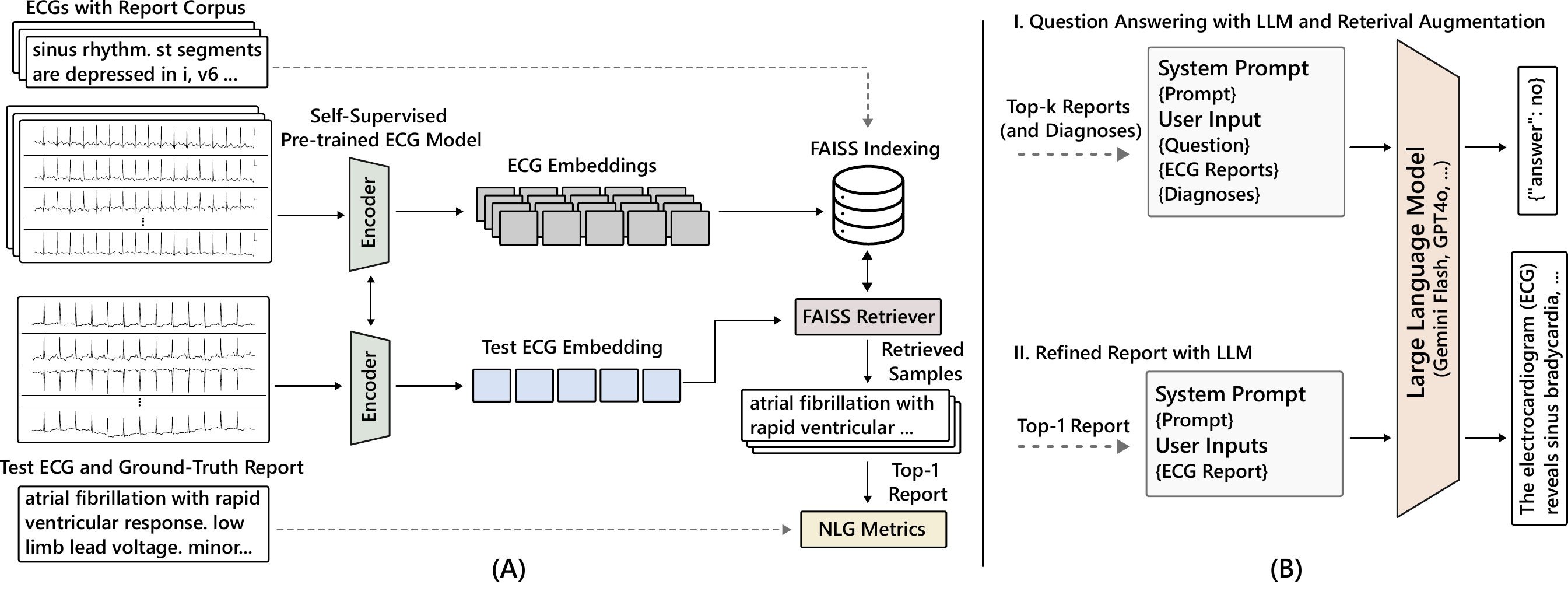}
\caption{Overview of \ours~leveraging self-supervised model to tackle report generation via retrieval and QA with LLMs. NLG refers to natural language generation.} 
\label{fig:spalsh}
\end{figure*}

To address these challenges, we propose a novel approach that pre-trains an ECG encoder to learn generalizable ECG embeddings for efficient similarity searches. Our encoder is trained in a self-supervised learning manner that integrates ECG signals with their corresponding textual reports. During inference, we use this encoder to find relevant reports based on the nearest neighbors of an ECG embedding. Specifically, we choose the top-$1$ most similar report as the generated report for a given ECG. This retrieval-based method forms the foundation for our question-answering system. We feed top-$k$ retrieved reports, along with their diagnoses labels, into the LLM for zero-shot question answering. The LLM processes this information to answer queries, utilizing the retrieved reports to provide more accurate and contextually relevant responses. Our approach combines the strengths of similarity-based retrieval and the natural language understanding capabilities of LLMs, potentially offering grounded and interpretable results compared to traditional fully supervised methods.

We validate our approach using two prominent ECG datasets: PTB-XL~\cite{wagner2020ptb} and MIMIC-IV-ECG~\cite{gow2023mimic}. Our ECG encoder, pre-trained on PTB-XL, is evaluated for report generation on both datasets, assessing performance in in-domain and cross-domain scenarios. The retrieval-based method demonstrates superior performance in both settings, underlining its robustness and generalizability. Additionally, we employ off-the-shelf LLMs for zero-shot question answering, achieving competitive results compared to fully supervised approaches without task-specific fine-tuning. This comprehensive evaluation showcases the effectiveness of our method, which combines self-supervised pre-training, dynamic retrieval, and LLM-based refinement to enhance automated ECG interpretation in clinical settings. 

\section{Method}
\label{sec:method}
We frame ECG-to-text report generation and question answering as a retrieval augmented generation task using a report corpus $R$. Our approach combines a self-supervised ECG model with efficient similarity search to generate reports and answer ECG-related questions. The method consists of four stages: (1) self-supervised pre-training, (2) embedding generation and indexing, (3) report retrieval and refinement, and (4) zero-shot LLM-based question-answering. Figure~\ref{fig:spalsh} illustrates our proposed technique.

We pre-train a multi-modal model to learn joint representations of ECG signals and their textual reports. Let $X = {x_1, x_2, ..., x_n}$ be a set of ECGs, where $x_i \in \mathbb{R}^{L \times C}$ represents an ECG with $L$ time steps and $C$ channels, which correspond to the leads. The corresponding reports are denoted as $R = {r_1, r_2, ..., r_n}$. We use a masked autoencoder-based self-supervised learning approach similar to~\cite{chen2022multi}, combining three loss terms: masked language modeling (MLM), masked ECG modeling (MEM), and ECG-text matching (ETM), as $L = L_{\text{MLM}} + L_{\text{MEM}} + L_{\text{ETM}}$. The MEM task involves masking a high proportion of ECG signal patches and reconstructing them. For MLM, we mask and predict a smaller proportion of tokens in the textual reports. The ETM loss uses a binary classifier to determine if ECG-text pairs are semantically aligned (positive) or misaligned (negative). We employ transformer-based architectures for both ECG and text encoders.

Our multi-modal Transformer-based architecture~\cite{NIPS2017_3f5ee243} for ECGs and texts comprises: 1) separate uni-modal encoders, 2) a multi-modal fusion module, and 3) separate uni-modal decoders for pre-training tasks. The text encoder uses a pre-trained BERT model~\cite{devlin2018bert}. The ECG encoder applies 1D convolutional layers to extract local features from signal before feeding them into a transformer. We follow the configurations from~\cite{chen2022multi} for other architectural aspects. The model is pre-trained using above defined MLM, MEM, and ETM tasks.

After pre-training, we generate embeddings for all ECG samples using the ECG encoder $f(\cdot)$: $z_i = f(x_i), z_i \in \mathbb{R}^d$, where $d$ is the embedding dimension. We build a FAISS~\cite{douze2024faiss} index for efficient similarity search (Figure~\ref{fig:spalsh}.A). During inference, for a new ECG $x_t$ sample, we compute $z_t = f(x_t)$ and retrieve $k$ nearest neighbors using FAISS. Let $N_k(z_t) = {(i_j, d_j) | j = 1, ..., k}$ denote the set of $k$ nearest neighbors, where $i_j$ is the index of the $j$-th neighbor and $d_j$ is the distance. We then retrieve the corresponding reports: $R_{\text{retrieved}} = {r_{i_j} | (i_j, d_j) \in N_k(z_t)}$.

\begin{table*}[t]
\centering
\caption{Performance comparison of various methods for ECG report generation on PTB-XL and MIMIC-IV-ECG datasets.}
\label{tab:report}
\resizebox{\textwidth}{!}{
\begin{tabular}{@{}llccccccc@{}}
\toprule
\textbf{Dataset} & \textbf{Method} & \textbf{BLEU-1} & \textbf{BLEU-2} & \textbf{BLEU-3} & \textbf{BLEU-4} & \textbf{BERTScore} & \textbf{Meteor} & \textbf{Rouge} \\ \midrule
\multicolumn{1}{l}{\multirow{4}{*}{\begin{tabular}[c]{@{}l@{}}PTB-XL\\ (in-domain)\end{tabular}}} & Random & 0.152 & 0.091 & 0.060 & 0.049 & 0.582 & 0.226 & 0.234 \\
\multicolumn{1}{l}{} & Common & 0.212 & 0.150 & 0.134 & 0.132 & 0.663 & 0.317 & 0.385 \\
\multicolumn{1}{l}{} & Transformer & 0.273 & 0.179 & 0.119 & 0.093 & 0.626 & 0.292 & 0.303 \\
\multicolumn{1}{l}{} & R2GenCMN (R)~\cite{chen2021cross} & 0.337 & 0.249 & 0.188 & 0.159 & 0.687 & 0.362 & 0.392 \\
\multicolumn{1}{l}{} & R2GenCMN (P)~\cite{chen2021cross} & 0.393 & 0.298 & 0.232 & 0.196 & 0.710 & 0.412 & 0.443 \\ \cmidrule(l){2-9} 
\multicolumn{1}{l}{} & \ours~(Ours) & \textbf{0.801} & \textbf{0.768} & \textbf{0.737} & \textbf{0.700} & \textbf{0.920} & \textbf{0.836} & \textbf{0.836} \\ \cmidrule(l){2-9} 
\multicolumn{1}{l}{\multirow{4}{*}{\begin{tabular}[c]{@{}l@{}}MIMIC-IV-ECG\\ (cross-domain)\end{tabular}}} & Random & 0.113 & 0.055 & 0.031 & 0.023 & 0.579 & 0.185 & 0.201 \\
\multicolumn{1}{l}{} & Common & 0.127 & 0.048 & 0.038 & 0.033 & 0.651 & 0.283 & 0.399 \\
\multicolumn{1}{l}{} & Transformer & 0.229 & 0.137 & 0.082 & 0.054 & 0.598 & 0.250 & 0.240 \\
\multicolumn{1}{l}{} & R2GenCMN (R)~\cite{chen2021cross} & 0.305 & 0.215 & 0.133 & 0.084 & 0.677 & 0.346 & 0.378 \\
\multicolumn{1}{l}{} & R2GenCMN (P)~\cite{chen2021cross} & 0.325 & 0.224 & 0.141 & 0.091 & 0.672 & 0.360 & 0.383 \\ \cmidrule(l){2-9} 
 & \ours~(Ours) & \textbf{0.348} & \textbf{0.271} & \textbf{0.212} & \textbf{0.182} & \textbf{0.714} & \textbf{0.419} & \textbf{0.439} \\ \bottomrule
\end{tabular}
}
\end{table*}

For a test ECG embedding $e_t$, we assign the closest retrieved sample's report as an initial prediction, that can be optionally refined by an LLM. For zero-shot ECG question answering (Figure~\ref{fig:spalsh}.B), we use the $k$ retrieved reports and their diagnoses labels as a way to perform in-context learning. We concatenate these as: $r_{\text{concat}} = r_{i_1} \oplus l_{i_1} \oplus r_{i_2} \oplus l_{i_2} \oplus \cdots \oplus r_{i_k} \oplus l_{i_k}$, where $r_{i_j} \in R_{\text{retrieved}}$, $l_{i_j}$ is the $j$-th report's diagnoses label, and $\oplus$ denotes concatenation. This concatenated input, the question, and a system prompt are passed to the LLM, instructing it to leverage the provided data to answer questions about the test ECG in a zero-shot manner. 

For pretraining, we use the PTB-XL dataset~\cite{wagner2020ptb} with 75\% and 15\% masking for MEM and MLM tasks, respectively. We use a batch size of $32$ and a learning rate of $5\times10^{-5}$, keeping other hyperparameters consistent with~\cite{chen2022multi}. We apply global max pooling over ECG encoder's output to get fixed-dimensional embeddings ($z$, $d = 768$). During indexing and retrieval, embeddings are L2 normalized. We set $k=1$ for report generation and $k=3$ for question-answering. Our ECG-QA LLM leverages GPT-4o (mini)\cite{openai2024gpt4omini}, Gemini-Flash1.5\cite{reid2024gemini}, and Llama3-70B~\cite{dubey2024llama} due to their cost-effectiveness. We set $\text{temperature}=1$ and $\text{max_tokens}=256$. Prompt details are provided in~\ref{box:prompt}. For \textit{question prompts}, we use variations for different answer formats: "Answer should be only yes and no" (single-verify), "Choose correct answer from the options provided below. If none... answer will be 'none'. If both conditions are present then provide both options as an answer." (single-choose), and a similar prompt for single-query allowing multiple answer selections. Answer options are always provided within the prompt.

\section{Experiments}  
\label{sec:experiments}
\noindent \textbf{Datasets.} We evaluate our report generation method on the PTB-XL~\cite{wagner2020ptb} and MIMIC-IV-ECG~\cite{gow2023mimic} datasets, both of which provide 12-lead ECG signals and corresponding textual reports. PTB-XL offers extensive cardiologist-annotated auxiliary information, including interpretive summaries, diagnostic assessments, likelihood estimates, and signal characteristics. MIMIC-IV-ECG provides machine-generated reports covering various conditions, which we combine into a single unified report. From this dataset, we randomly select $5$k samples from the test set for cross-domain evaluation. For question answering, we utilize the ECG-QA~\cite{oh2024ecg} dataset, which comprises curated questions about key ECG aspects. We experiment with verify, choose, and query question types, utilizing the provided train, test, and validation splits based on patient IDs to ensure no overlap among sets.\\

\begin{tcolorbox}[
    colframe=black,
    title=\footnotesize{LLM System Prompt for \textit{Zero-Shot} ECG-QA Task.}
]
Given the closest ECG retrieved reports and diagnoses to the test ECG as discovered by a multimodal model. Your job is to analyze the report and only answer the question. Output should be JSON of the following structure: \{`answer': ...\}. 
\newline
Question Specific Prompt: \texttt{\$\{question_prompt\}}
\newline 
Think step-by-step to generate an answer without any explanation.
\newline
ECG Reports: \texttt{\$\{reports\}}\newline
Diagnoses: \texttt{\$\{diagnoses\}}\newline
Question: \texttt{\$\{question\}}
\label{box:prompt}
\end{tcolorbox} 

\noindent \textbf{Baseline and Evaluation Metrics.} We conduct a quantitative evaluation of our approach with commonly-used metrics concerning both language fluency and accuracy. For report generation task, we compute several natural language generation (NLG) metrics, including BLEU-1,2,3 and 4~\cite{papineni2002bleu}, BERTScore~\cite{zhang2019bertscore}, Meteor~\cite{denkowski2011meteor} and Rouge~\cite{lin2004rouge}. These metrics cover various aspects of lexical and semantic similarities. We compare our method for this task with multiple baselines, a) randomly selected report, b) most common report, and c) R2GenCMN~\cite{chen2020generating} that we adapt for ECG signals, where we use both randomly initialized (R) and pre-trained (P) ECG encoders. For zero-shot ECG-QA task, we compute exact match accuracy to compare with prior methods and compare against six baselines~\cite{oh2024ecg} that are supervised in nature, where models are trained for this task. \\

\begin{figure*}[t]
\centering\includegraphics[width=\textwidth]{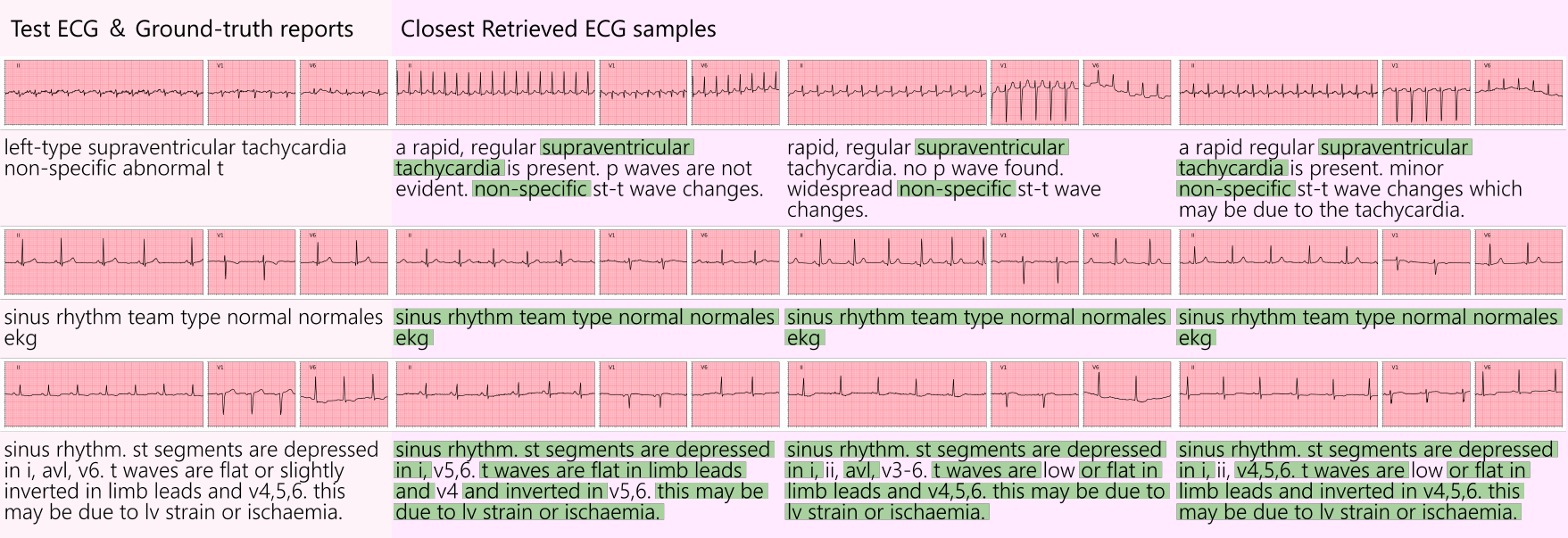}
\caption{Qualitative analysis of ground-truth and the top-$3$ retrieved reports along with their ECG signals on the PTB-XL dataset.} 
\label{fig:gen_rpt_ptbxl}
\end{figure*}

\begin{table}[htbp]
\centering
\caption{Evaluation of~\ours~paired with an off-the-shelf LLM on the ECG-QA dataset against supervised approaches. `S' refers to Single.}
\label{tab:qa}
\resizebox{\columnwidth}{!}{%
\begin{tabular}{@{}llccc@{}}
\toprule
\textbf{Method}   & \textbf{Model}     & \textbf{S-Verify} & \textbf{S-Choose} & \textbf{S-Query} \\ \midrule
per Q-type majority                   & -                  &  67.7             &  31.2               &  23.2              \\ \midrule
\multirow{5}{*}{Supervised~\cite{oh2024ecg}}           & M3AE            &  74.6             &  57.1               &  41.0              \\
& MedViLL            &  73.9             &  54.1               &  40.4              \\
& Fusion Transformer &  72.1             &  46.4               &  37.4              \\
& Blind Transformer  &  67.7             &  31.0               &  24.0              \\
& Deaf Transformer   &  67.3             &  31.4               &  27.0              \\ \midrule
\multirow{3}{*}{\ours~(Ours)} & Gemini Flash 1.5   &   72.54              &    58.52             &     32.57           \\
& GPT-4o mini         &   72.79              &     58.66            &     30.56           \\
& Llama3-70B         &    71.99             &      54.83           &    32.02            \\ \bottomrule
\end{tabular}
}
\end{table}

\noindent \textbf{Results.}
The main experimental results on report generation task are presented in Table~\ref{tab:report}. The results demonstrate the effectiveness of~\ours~across all evaluated metrics. On the PTB-XL dataset, it achieves substantial improvements over the R2GenCMN technique with pre-trained model, with a BLEU-4 score of $0.700$ compared to $0.196$, and a BERTScore of $0.920$ versus $0.710$. This significant performance gain highlights the superiority of our retrieval-augmented approach in capturing and generating accurate ECG reports. The cross-domain evaluation on MIMIC-IV-ECG shows that our method maintains its superior performance, albeit with a smaller margin, achieving a BLEU-4 score of $0.182$ and a BERTScore of $0.714$. This robustness in cross-domain scenarios underscores the generalizability of our method. 

Notably,~\ours~consistently outperforms both random and common baselines, as well as the R2GenCMN variants, across all metrics on both datasets. The high ROUGE and METEOR scores further indicate that our method generates reports with better content coverage and semantic similarity to ground truth reports. Further, Figure~\ref{fig:gen_rpt_ptbxl} shows qualitative results of retrieving similar examples with reports that closely match the ground truth, capturing key diagnostic features. These results collectively demonstrate the efficacy and simplicity of leveraging self-supervised representations for retrieval-based report generation. Our approach not only produces high-quality reports but also enables transparency by allowing clinicians to inspect and compare the generated report with similar examples. 

Table~\ref{tab:qa} presents the performance of various methods on three ECG question-answering tasks. Our proposed ECG-ReGen-QA approach, operating in a zero-shot setting, demonstrates competitive performance compared to supervised methods. Notably, our method paired with Gemini Flash 1.5 and GPT-4o mini achieves the highest scores on the S-Choose task ($58.52$\% and $58.66$\%, respectively), surpassing all supervised models. For the S-Verify task, our approach performs comparably to the best supervised model (M3AE), with scores ranging from $71.99$\% to $72.79$\%. While ECG-ReGen-QA shows lower performance on the S-Query task (open-ended questions and answers) compared to the top supervised models, it still significantly outperforms the per Q-type majority baseline and other supervised models like the Blind and Deaf Transformers. These results are particularly impressive considering that our method operates in a zero-shot manner, requiring no task-specific fine-tuning. This demonstrates the effectiveness of our retrieval-augmented approach in leveraging off-the-shelf language models for ECG analysis. Likewise, the performance gap between different considered language models is relatively small, suggesting that the choice of LLM may not be crucial given that retrieved samples with self-supervised ECG model are similar to the test case. 

\section{Conclusions}  
\label{sec:conclusions}
This work introduces a novel retrieval-based method for ECG report generation and question answering, leveraging self-supervised pre-training, efficient similarity search, and LLM-powered zero-shot question answering. Our approach demonstrates superior performance in both in-domain and cross-domain evaluations for report generation task, showcasing improved efficiency, inherent explainability, and enhanced generalizability. By integrating LLMs for zero-shot QA, we further augment the system's capabilities, offering a promising avenue for accurate ECG interpretation with potential benefits for cardiologist workflow and patient care.

\bibliographystyle{IEEEbib}
\bibliography{main}

\begin{thebibliography}{10}

\bibitem{minchole2019artificial}
Ana Minchol{\'e} and Blanca Rodriguez,
\newblock ``Artificial intelligence for the electrocardiogram,''
\newblock {\em Nature medicine}, vol. 25, no. 1, pp. 22--23, 2019.

\bibitem{moor2023foundation}
Michael Moor, Oishi Banerjee, Zahra Shakeri~Hossein Abad, Harlan~M Krumholz, Jure Leskovec, Eric~J Topol, and Pranav Rajpurkar,
\newblock ``Foundation models for generalist medical artificial intelligence,''
\newblock {\em Nature}, vol. 616, no. 7956, pp. 259--265, 2023.

\bibitem{chen2020generating}
Zhihong Chen, Yan Song, Tsung-Hui Chang, and Xiang Wan,
\newblock ``Generating radiology reports via memory-driven transformer,''
\newblock {\em arXiv preprint arXiv:2010.16056}, 2020.

\bibitem{yan2023style}
Benjamin Yan, Ruochen Liu, David~E Kuo, Subathra Adithan, Eduardo~Pontes Reis, Stephen Kwak, Vasantha~Kumar Venugopal, Chloe~P O'Connell, Agustina Saenz, Pranav Rajpurkar, et~al.,
\newblock ``Style-aware radiology report generation with radgraph and few-shot prompting,''
\newblock {\em arXiv preprint arXiv:2310.17811}, 2023.

\bibitem{thirunavukarasu2023large}
Arun~James Thirunavukarasu, Darren Shu~Jeng Ting, Kabilan Elangovan, Laura Gutierrez, Ting~Fang Tan, and Daniel Shu~Wei Ting,
\newblock ``Large language models in medicine,''
\newblock {\em Nature medicine}, vol. 29, no. 8, pp. 1930--1940, 2023.

\bibitem{oh2024ecg}
Jungwoo Oh, Gyubok Lee, Seongsu Bae, Joon-myoung Kwon, and Edward Choi,
\newblock ``Ecg-qa: A comprehensive question answering dataset combined with electrocardiogram,''
\newblock {\em Advances in Neural Information Processing Systems}, vol. 36, 2024.

\bibitem{wan2024electrocardiogram}
Zhongwei Wan, Che Liu, Xin Wang, Chaofan Tao, Hui Shen, Zhenwu Peng, Jie Fu, Rossella Arcucci, Huaxiu Yao, and Mi~Zhang,
\newblock ``Electrocardiogram instruction tuning for report generation,''
\newblock {\em arXiv preprint arXiv:2403.04945}, 2024.

\bibitem{endo2021retrieval}
Mark Endo, Rayan Krishnan, Viswesh Krishna, Andrew~Y Ng, and Pranav Rajpurkar,
\newblock ``Retrieval-based chest x-ray report generation using a pre-trained contrastive language-image model,''
\newblock in {\em Machine Learning for Health}. PMLR, 2021, pp. 209--219.

\bibitem{wagner2020ptb}
Patrick Wagner, Nils Strodthoff, Ralf-Dieter Bousseljot, Dieter Kreiseler, Fatima~I Lunze, Wojciech Samek, and Tobias Schaeffter,
\newblock ``Ptb-xl, a large publicly available electrocardiography dataset,''
\newblock {\em Scientific data}, vol. 7, no. 1, pp. 1--15, 2020.

\bibitem{gow2023mimic}
Brian Gow, Tom Pollard, Larry~A Nathanson, Alistair Johnson, Benjamin Moody, Chrystinne Fernandes, Nathaniel Greenbaum, Seth Berkowitz, Dana Moukheiber, Parastou Eslami, et~al.,
\newblock ``Mimic-iv-ecg-diagnostic electrocardiogram matched subset,''
\newblock {\em Type: dataset}, 2023.

\bibitem{chen2022multi}
Zhihong Chen, Yuhao Du, Jinpeng Hu, Yang Liu, Guanbin Li, Xiang Wan, and Tsung-Hui Chang,
\newblock ``Multi-modal masked autoencoders for medical vision-and-language pre-training,''
\newblock in {\em International Conference on Medical Image Computing and Computer-Assisted Intervention}. Springer, 2022, pp. 679--689.

\bibitem{NIPS2017_3f5ee243}
Ashish Vaswani, Noam Shazeer, Niki Parmar, Jakob Uszkoreit, Llion Jones, Aidan~N Gomez, \L~ukasz Kaiser, and Illia Polosukhin,
\newblock ``Attention is all you need,''
\newblock in {\em Advances in Neural Information Processing Systems}, I.~Guyon, U.~Von Luxburg, S.~Bengio, H.~Wallach, R.~Fergus, S.~Vishwanathan, and R.~Garnett, Eds. 2017, vol.~30, Curran Associates, Inc.

\bibitem{devlin2018bert}
Jacob Devlin,
\newblock ``Bert: Pre-training of deep bidirectional transformers for language understanding,''
\newblock {\em arXiv preprint arXiv:1810.04805}, 2018.

\bibitem{douze2024faiss}
Matthijs Douze, Alexandr Guzhva, Chengqi Deng, Jeff Johnson, Gergely Szilvasy, Pierre-Emmanuel Mazar{\'e}, Maria Lomeli, Lucas Hosseini, and Herv{\'e} J{\'e}gou,
\newblock ``The faiss library,''
\newblock {\em arXiv preprint arXiv:2401.08281}, 2024.

\bibitem{chen2021cross}
Zhihong Chen, Yaling Shen, Yan Song, and Xiang Wan,
\newblock ``Cross-modal memory networks for radiology report generation,''
\newblock in {\em Proceedings of the 59th Annual Meeting of the Association for Computational Linguistics and the 11th International Joint Conference on Natural Language Processing (Volume 1: Long Papers)}, 2021, pp. 5904--5914.

\bibitem{openai2024gpt4omini}
OpenAI,
\newblock ``Gpt-4o (mini),'' \texttt{https://openai.com}, 2024,
\newblock [Large language model].

\bibitem{reid2024gemini}
Machel Reid, Nikolay Savinov, Denis Teplyashin, Dmitry Lepikhin, Timothy Lillicrap, Jean-baptiste Alayrac, Radu Soricut, Angeliki Lazaridou, Orhan Firat, Julian Schrittwieser, et~al.,
\newblock ``Gemini 1.5: Unlocking multimodal understanding across millions of tokens of context,''
\newblock {\em arXiv preprint arXiv:2403.05530}, 2024.

\bibitem{dubey2024llama}
Abhimanyu Dubey, Abhinav Jauhri, Abhinav Pandey, Abhishek Kadian, Ahmad Al-Dahle, Aiesha Letman, Akhil Mathur, Alan Schelten, Amy Yang, Angela Fan, et~al.,
\newblock ``The llama 3 herd of models,''
\newblock {\em arXiv preprint arXiv:2407.21783}, 2024.

\bibitem{papineni2002bleu}
Kishore Papineni, Salim Roukos, Todd Ward, and Wei-Jing Zhu,
\newblock ``Bleu: a method for automatic evaluation of machine translation,''
\newblock in {\em Proceedings of the 40th annual meeting of the Association for Computational Linguistics}, 2002, pp. 311--318.

\bibitem{zhang2019bertscore}
Tianyi Zhang, Varsha Kishore, Felix Wu, Kilian~Q Weinberger, and Yoav Artzi,
\newblock ``Bertscore: Evaluating text generation with bert,''
\newblock {\em arXiv preprint arXiv:1904.09675}, 2019.

\bibitem{denkowski2011meteor}
Michael Denkowski and Alon Lavie,
\newblock ``Meteor 1.3: Automatic metric for reliable optimization and evaluation of machine translation systems,''
\newblock in {\em Proceedings of the sixth workshop on statistical machine translation}, 2011, pp. 85--91.

\bibitem{lin2004rouge}
Chin-Yew Lin,
\newblock ``Rouge: A package for automatic evaluation of summaries,''
\newblock in {\em Text summarization branches out}, 2004, pp. 74--81.

\end{thebibliography}
\end{document}